\documentclass[%
aip,
 amsmath,amssymb,
 reprint,%
]{revtex4-1}

\usepackage{graphicx}% Include figure files
\usepackage{dcolumn}% Align table columns on decimal point
\usepackage{bm}% bold math
\usepackage[caption=false]{subfig}

\usepackage[utf8]{inputenc}
\usepackage[T1]{fontenc}
\usepackage{mathptmx}
\usepackage{etoolbox}
\usepackage{xcolor}

\usepackage{hyperref}

\makeatletter
\def\@email#1#2{%
 \endgroup
 \patchcmd{\titleblock@produce}
  {\frontmatter@RRAPformat}
  {\frontmatter@RRAPformat{\produce@RRAP{*#1\href{mailto:#2}{#2}}}\frontmatter@RRAPformat}
  {}{}
}%
\makeatother
\begin{document}

\preprint{AIP/123-QED}

\title[Discovery of sparse hysteresis models for piezoelectric materials]{Discovery of sparse hysteresis models for piezoelectric materials}
% Force line breaks with \\
\author{Abhishek Chandra}
 \email{Authors to whom correspondence should be addressed: Abhishek Chandra (a.chandra@tue.nl)}
 \email{a.chandra@tue.nl, b.daniels@tue.nl, m.curti@tue.nl, k.tiels@tue.nl, e.lomonova@tue.nl, tartakovsky@stanford.edu}
 %\altaffiliation{The article has been submitted to an AIP journal}%Lines break automatically or can be forced with \\
\author{Bram Daniels}%
\affiliation{ 
Electromechanics and Power Electronics Group, Department of Electrical Engineering, Eindhoven University of Technology, 5600 MB Eindhoven, The Netherlands
}%
\author{Mitrofan Curti}%
\affiliation{ 
Electromechanics and Power Electronics Group, Department of Electrical Engineering, Eindhoven University of Technology, 5600 MB Eindhoven, The Netherlands
}%

\author{Koen Tiels}
 %\homepage{http://www.Second.institution.edu/~Charlie.Author.}
\affiliation{%
Control Systems Technology Group, Department of Mechanical Engineering, Eindhoven University of Technology, 5600 MB Eindhoven, The Netherlands%\\This line break forced% with \\
}%

\author{Elena A. Lomonova}
\affiliation{
Electromechanics and Power Electronics Group, Department of Electrical Engineering, Eindhoven University of Technology, 5600 MB Eindhoven, The Netherlands
}

\author{Daniel M. Tartakovsky}%
\affiliation{ 
Department of Energy Science and Engineering, Stanford University, 367 Panama Street, Stanford,  California  94305, United States}%

\date{\today}% It is always \today, today,

\begin{abstract}
This article presents an approach for modelling hysteresis in piezoelectric materials, that leverages recent advancements in machine learning, particularly in sparse-regression techniques. While sparse regression has previously been used to model various scientific and engineering phenomena, its application to nonlinear hysteresis modelling in piezoelectric materials has yet to be explored. The study employs the least-squares algorithm with a sequential threshold to model the dynamic system responsible for hysteresis, resulting in a concise model that accurately predicts hysteresis for both simulated and experimental piezoelectric material data. Several numerical experiments are performed, including learning butterfly-shaped hysteresis and modelling real-world hysteresis data for a piezoelectric actuator. The presented approach is compared to traditional regression-based and neural network methods, demonstrating its efficiency and robustness. Source code is available at \href{https://github.com/chandratue/SmartHysteresis}{https://github.com/chandratue/SmartHysteresis.}
\end{abstract}

\maketitle

%\begin{quotation}
%The ``lead paragraph'' is encapsulated with the \LaTeX\ 
%\verb+quotation+ environment and is formatted as a single paragraph before %the first section heading. 
%(The \verb+quotation+ environment reverts to its usual meaning after the first sectioning command.) 
%Note that numbered references are allowed in the lead paragraph.
%
%The lead paragraph will only be found in an article being prepared for the journal \textit{Chaos}.
%\end{quotation}

%\section{\label{introduction}Introduction}

Constitutive relationships are unique to specific materials and cannot be universally applied across different materials. \cite{ottosen2005mechanics, bertotti1998hysteresis} Despite this, these materials' properties and characteristics, such as hysteresis, can be similar. For example, hysteresis in piezoelectric materials (piezoelectric actuators) used in micro and nano-positioning devices can reduce their positioning precision. \cite{lin2012tracking, gan2019review} Accurate calculations of the electric field-strain relationship between voltage and displacement require a constitutive relationship specific to the material.

In practice, constitutive relationships exhibit nonlinear hysteresis behaviour, and accurate hysteresis models are crucial for tasks such as high-resolution positioning. Conventional methods for modelling hysteresis in piezoelectric materials include experimental or phenomenological models based on established theories. Examples of phenomenological models include the Duhem model, Bouc-Wen model, Jiles-Atherton model and Preisach model. \cite{bertotti2005science} While these models have proven effective, they also have limitations. For example, the Preisach model requires a weight function often estimated through empirical methods or polynomial and spline fitting, adding complexity to the modelling process. \cite{zeinali2019comparison, 10034669} Additionally, the Preisach model does not provide physical insights into the underlying mechanisms of hysteresis, making it difficult to interpret the results and make predictions for new situations.\cite{bertotti1998hysteresis}

With the availability of increased measurement data, several data-driven approaches for modelling and predicting hysteresis have emerged. Deep Neural Networks (DNNs), known for their exceptional performance in data-science applications, have been utilized to model hysteresis. \cite{kim2019response, wei2000constructing} Although DNNs excel at prediction, they lack interpretability due to the absence of explicit governing equations.\cite{montans2019data} This drawback can be overcome by evolutionary algorithms, which may uncover the structure of nonlinear dynamic systems from data and are thus a viable alternative for modelling hysteresis.\cite{kyprianou2001identification} Despite this, their implementation is computationally intensive and difficult to scale to larger situations. To address this, machine learning algorithms that promote sparsity offer a computationally efficient option, blending the capabilities of symbolic regression with computational feasibility. Consequently, such techniques may result in parsimonious and interpretable models. \cite{montans2019data}

We build on the work by Chandra et al. \cite{chandra2022data} and present a white-box technique to model hysteresis in piezoelectric materials using the Sparse Identification of Nonlinear Dynamics (SINDy) framework. This framework, introduced in \cite{brunton2016discovering} has demonstrated its potential in learning nonlinear dynamical systems. \cite{lai2019sparse} The proposed technique does not require assumptions such as rate dependency, monotonicity, and congruency during the modelling phase. However, the resulting model exhibits these properties for a piezoelectric material due to the accurate input-output mapping. As a result, the white-box model makes efficient system analysis possible for piezoelectric materials. 

The major contributions of the manuscript are as follows,

\begin{itemize}
    \item A sparse regression-based methodology for modelling hysteresis in piezoelectric materials is introduced.
    \item It is shown that the presented methodology could be used to model constitutive hysteresis relationships in a white-box dynamical form.
    \item Several numerical experiments are presented for piezoelectric datasets and compared with traditional regression and neural network algorithms.
\end{itemize}

%The rest of the article is organized as follows. Section II introduces the methodology to model scalar hysteresis as a dynamical system. Section III presents the numerical experiments for piezoelectric actuators. Section IV contains some recommendations and insights about employing the sparse-regression framework for magnetic materials. Finally, conclusions are presented in section V. 

%\section{Methodology}
The structure of the letter is as follows: First, we present our methodology for modelling hysteresis as a sparse dynamical equation, leveraging the inherent sparsity and dynamics of hysteresis. To validate the effectiveness of our methodology, we conduct two numerical experiments employing the Duhem and Bouc-Wen models. These experiments serve as a proof of concept, demonstrating the capability of our approach to capture and replicate hysteresis behaviour accurately. Moreover, we conduct a comprehensive comparative analysis, contrasting our methodology with conventional regression-based and deep neural network approaches. The evaluation considers training data size, noisy data, and library selection.

Additionally, we apply our proposed methodology to model a real-world piezoelectric dataset, assessing the resulting model's accuracy and training time compared to previously published models. Lastly, we address the modelling of butterfly-shaped hysteresis data and present a straightforward extension of the Duhem model to encompass butterfly-type hysteresis. Through these investigations, we emphasize our approach's superior performance and versatility in accurately modelling hysteresis phenomena for piezoelectric materials. The letter concludes with a summary of findings and implications.

In the following, we detail the methodology employed to model sparse hysteresis relationships. To obtain a simple hysteresis model, the approach involves obtaining time-series data of input and output ($u \in \mathbb{R}^n$ and $w \in \mathbb{R}^n$), with time derivatives either measured directly or computed numerically. The data are then split into training and testing sets and organized into matrices $X$ and $\dot{X}$ of size $\mathbb{R}^{n \times 2}$, with $1\leq i\leq n$.

$ X = \begin{pmatrix}
\vline & \vline \\
w(t_i) & u(t_i) \\
\vline & \vline &
\end{pmatrix}$,
\quad
$ \dot{X} = \begin{pmatrix}
\vline & \vline \\
\dot{w}(t_i) & \dot{u}(t_i)\\
\vline & \vline & 
\end{pmatrix}$,
\\
where ($^.$) refers to the time derivative of the quantity. In order to obtain a simple model, a set of candidate functions ($\Theta \in \mathbb{R}^{n \times m}$) is selected, with the assumption that a sparse combination of these functions dictates the hysteresis dynamics. The columns of $\Theta$ represent the potential space of basis functions, which include nonlinear functions.

$ \Theta(X, \dot{X}) = \begin{pmatrix}
\vline & \vline & \vline & \vline & \vline \\
\theta_1 & \vline & \theta_j & \vline & \theta_m \\
\vline & \vline & \vline & \vline & \vline
\end{pmatrix}$,
\quad 
$ \Xi = \begin{pmatrix}
\vline & \vline \\
\xi_{1_m} & \xi_{2_m} \\
\vline & \vline
\end{pmatrix}$.

The objective is to obtain a set of sparse coefficient vectors $\Xi \in \mathbb{R}^{m \times 2}$ that represent the coefficients for the linear combination of basis functions in the library $\Theta$. The problem of approximating hysteresis dynamics can be expressed succinctly as
\begin{equation}
    \dot{X} = \Theta(X, \dot{X})\Xi.
\end{equation}

The study solves the hysteresis modelling problem using the Sequential Threshold Least-Squares (STLSQ) algorithm. \cite{brunton2019data} The STLSQ algorithm is applied to the approximation problem, which is defined using the time-series input and output data and a candidate library of basis functions. The algorithm returns a sparse coefficient vector $\Xi$. Most of the coefficients become zero, and the basis functions corresponding to the non-zero elements of $\Xi$ represent the sparse dynamic relationship that dictates the hysteresis behaviour. The accuracy of the obtained models is evaluated using the metric relative percent error. The library is prepared using terms from the Bouc-Wen hysteresis model, Duhem hysteresis model, and higher-order polynomials and their combinations. The threshold ($\lambda$) for the STLSQ algorithm is chosen based on the dataset to ensure that the obtained model is accurate and sparse.

\begin{figure}[htp]
\subfloat{\includegraphics[clip,width=0.75\columnwidth]{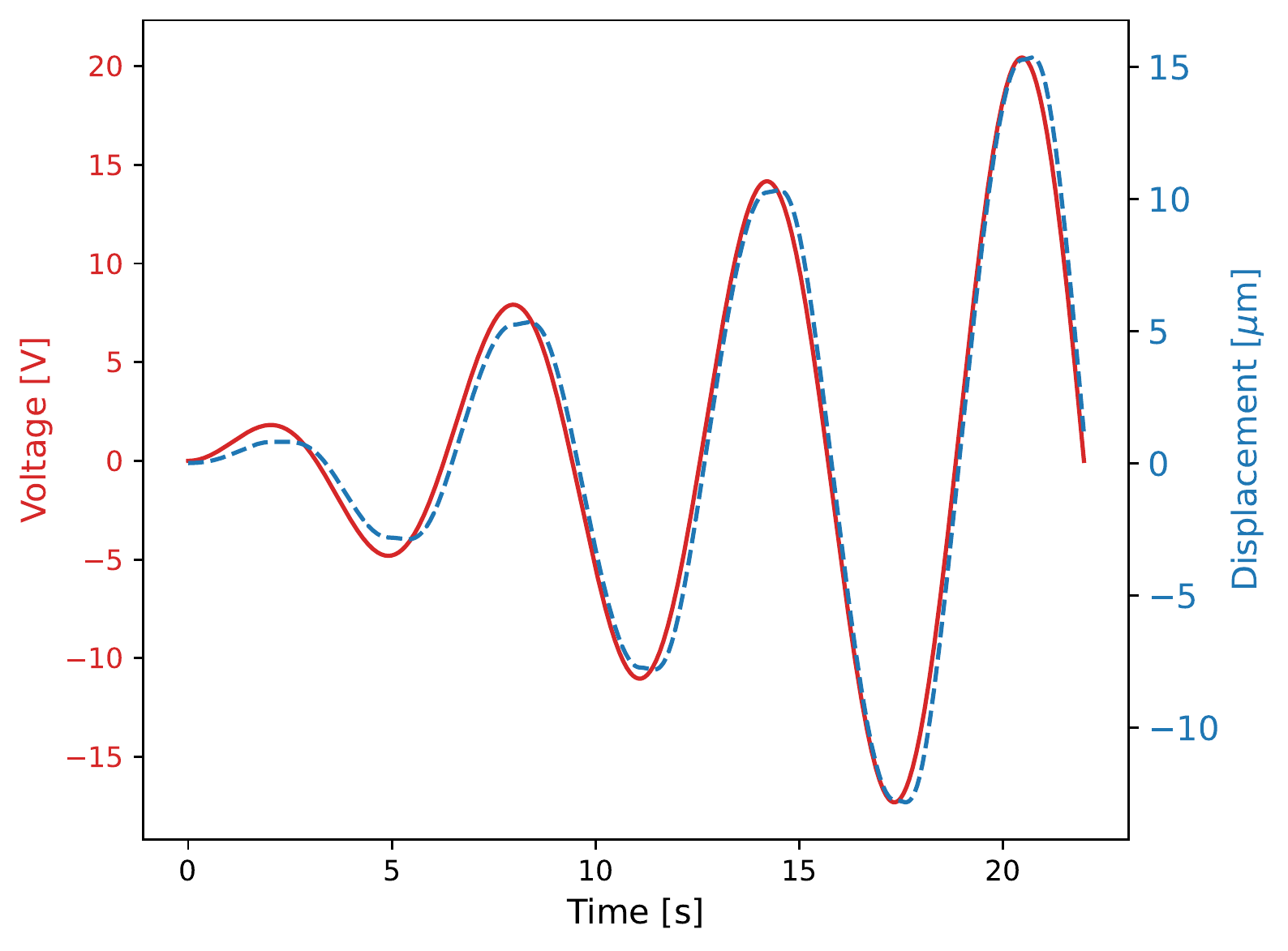}}

\subfloat{\includegraphics[clip,width=0.75\columnwidth]{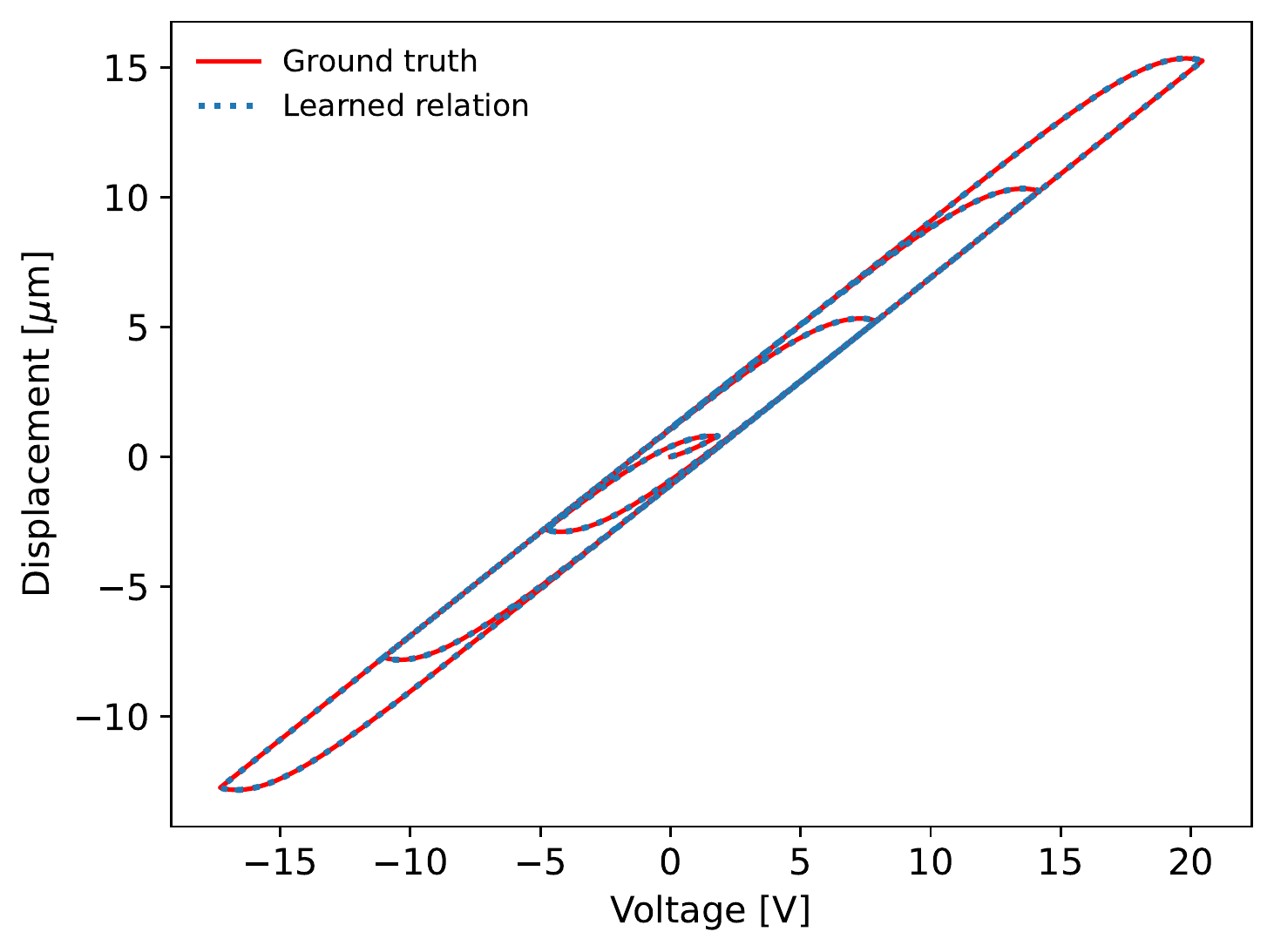}}
\caption{\textbf{Learning hysteresis model from the data generated by simulating Duhem model (top:)} Input voltage and output displacement varying with respect to time. \textbf{(bottom:)} Learned voltage vs displacement relationship compared against the simulated data acting as ground truth.}
\label{fig1}
\end{figure}

\begin{figure}[htp]
\subfloat{\includegraphics[clip,width=0.75\columnwidth]{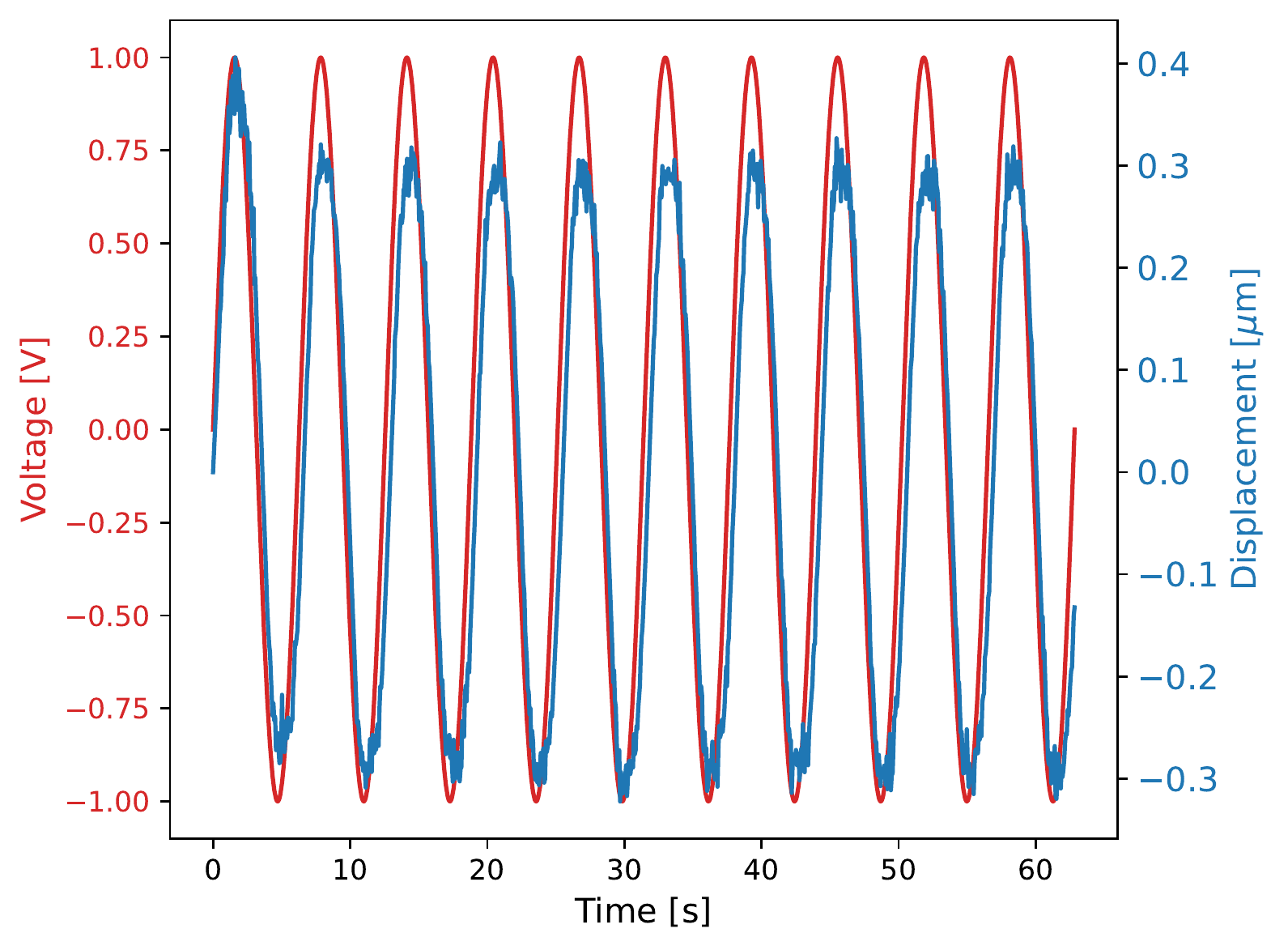}}

\subfloat{\includegraphics[clip,width=0.75\columnwidth]{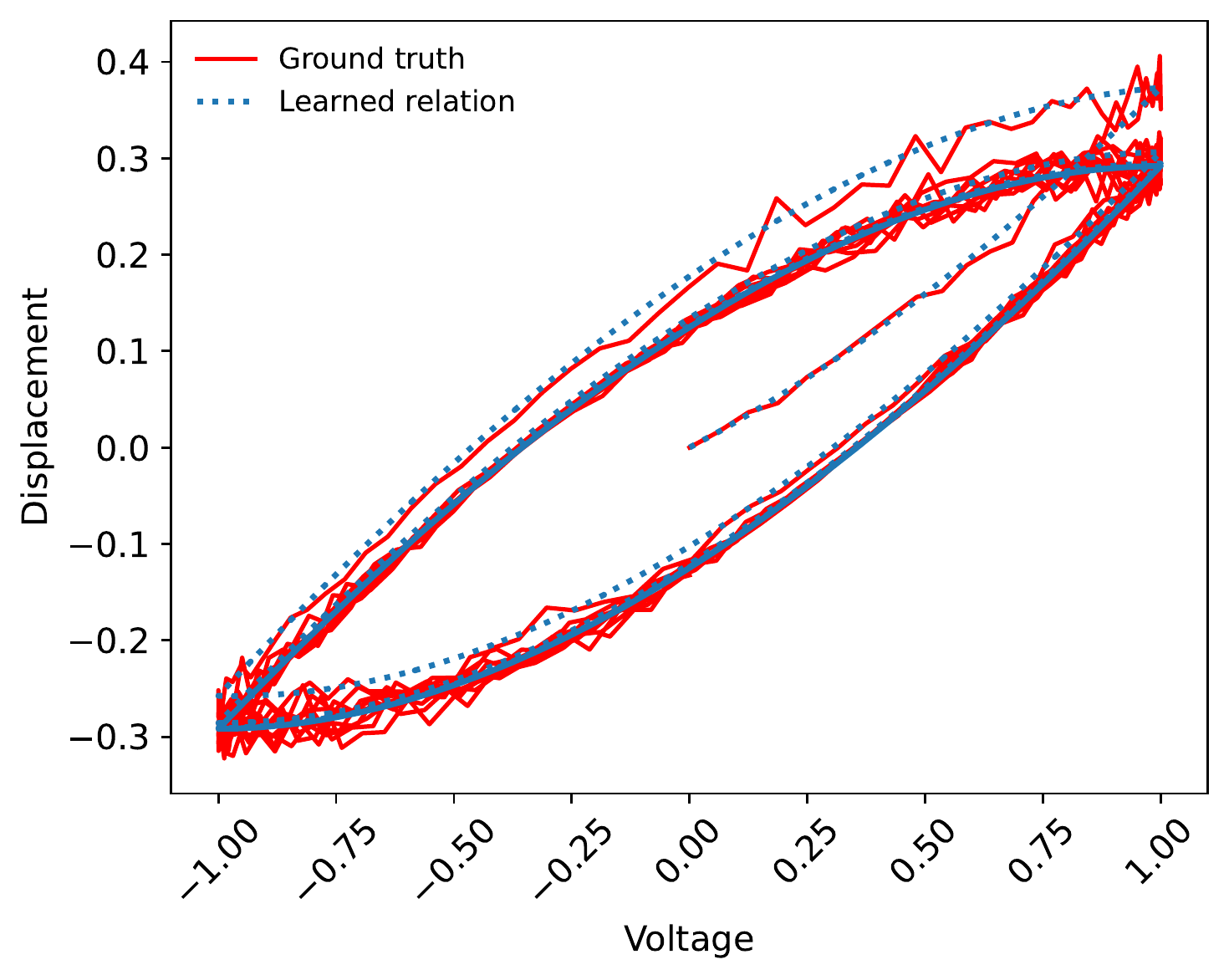}}
\caption{\textbf{Learning hysteresis model from the noisy data generated by corrupting the Bouc-Wen output with $5\%$ Gaussian noise (top:)} Input voltage and noisy output displacement varying with respect to time. \textbf{(bottom:)} Learned voltage vs displacement relationship compared against the noisy simulated data acting as ground truth.}
\label{fig2}
\end{figure}

%The next section presents the numerical experiments on the piezoelectric actuator's simulated and experimental data sets.

%\section{Numerical Experiments}

%\subsection{Simulated dataset}
Next, we present numerical experiments to validate the methodology. Two numerical experiments have been carried out on computer-simulated data to validate the method with threshold $\lambda = 0.1$. Source code is available at \href{https://github.com/chandratue/SmartHysteresis}{https://github.com/chandratue/SmartHysteresis.} Comprehensive instructions for accessing and executing the code are provided therein. For the first experiment, the data are produced utilizing the Duhem hysteresis model, which is appropriate for explaining hysteresis in piezoelectric materials. \cite{lin2012tracking} PySINDy,\cite{de2020pysindy} the SINDy library implemented in Python was used to perform time differentiation of input-output data. The Duhem model characterizes the relationship between the input and output as $\dot{w}(t) = \alpha |\dot{u}(t)| u(t) - \beta |\dot{u}(t)| w(t) + \gamma \dot{u}(t)$, where $\alpha$, $\beta$, and $\gamma$ are parameters controlling the hysteresis loop's shape. Taking $\alpha =0.4$, $\beta=0,5$, and $\gamma =0.25$ with input as presented in Fig. \ref{fig1}. Once the STLSQ algorithm is applied, the resulting hysteresis model is expressed as %$\dot{w}(t) = 0.99 |\dot{u}(t)| u(t) - 0.249 |\dot{u}(t)| w(t) + 19.99 \dot{u}(t)$
\begin{equation}
     \dot{w}(t) = 0.4 |\dot{u}(t)| u(t) - 0.5 |\dot{u}(t)| w(t) + 0.251 \dot{u}(t)
\end{equation}
accounting for $5.8 \times 10^{-6}\%$ error. The relative percentage error $\mathcal{R}$ is chosen for the error metric.
\begin{equation}
\begin{aligned}
\mathcal{R} = \frac{||w^{*} - w ||_{2}}{||w||_{2}} \times 100
\end{aligned}
\end{equation}
where $w^{*}$ is the simulated solution of the approximated model.

For the second experiment, the data are generated from the Bouc-Wen model \cite{ottosen2005mechanics} given by, ${\dot{w}(t) = \alpha \dot{u}(t) - \beta |\dot{u}(t)| |w(t)|^{n-1} w(t) - \gamma \dot{u}(t) |w(t)|^{n}}$,  with $\alpha =5$, $\beta=0.25$, $\gamma =0.5$, and $n=1$. The learned model has $\mathcal{R} = 4.6  \times 10^{-6}\%$ error and the learned model is%$\dot{w}(t) = 0.99 \dot{u}(t) - 0.5 |\dot{u}(t)| |w(t)|^{n-1} w(t) - 1.97 \dot{u}(t) |w(t)|^{n}$.
\begin{equation}
     \dot{w}(t) = 4.99 \dot{u}(t) - 0.25 |\dot{u}(t)| |w(t)|^{n-1} w(t) - 0.49 \dot{u}(t) |w(t)|^{n}
\end{equation}

%\subsection{Comparisons}
The presented experiments on the simulated dataset for the Bouc-Wen and Duhem models show that the method is applicable to a hysteresis dataset. However, given the input-output dataset, numerous traditional regression-based and neural network-based methods could be used to establish the underlying relationship. We compare our method with three traditional methods: linear regression, ridge regression and feedforward deep neural network. However, the methods mentioned above need a relatively smaller feature library. For the next two experiments, we simulate the Bouc-Wen model and use its features for these methods for a fair comparison.

First, we test the performance of sparse discovery when the number of data points is limited and generated from the Bouc-Wen model with parameters $\alpha =1$, $\beta=0.5$, $\gamma =2$, and $n=1$. We perform experiments with $1000$, $5000$, and $10000$ training points and observe that as the number of training points increases, the method performs better and even achieves a near-perfect model for the simulated smooth dataset, which other methods could not achieve, as presented in Table \ref{tab:table1}. Furthermore, for as less as $1000$ training points, SINDy outperforms the traditional methods and learns the relationship with the least error. 

Next, we test our method for a noisy dataset. The simulated dataset generated from the Bouc-Wen model with parameters $\alpha =0.4$, $\beta=0.5$, $\gamma =0.25$, is corrupted with $1$ percent and $5$ percent Gaussian noise, respectively. The results for SINDy and traditional methods are presented in Table \ref{tab:table2} and Fig. \ref{fig2}. $1$ percent Gaussian noise is common for raw experimental data under controlled conditions. Our method learns the model with $\mathcal{R} = 0.01$ percent accuracy by learning the exact form of the dynamical system with errors in the coefficients. Additionally, in the case of $5$ percent Gaussian noise, which could happen under less controlled experiments,\cite{lai2019sparse} the method still achieves less than $0.3$ percent accuracy and outperforms the traditional methods. 

\begin{figure}[!htbp]
\subfloat{\includegraphics[clip,width=0.75\columnwidth]{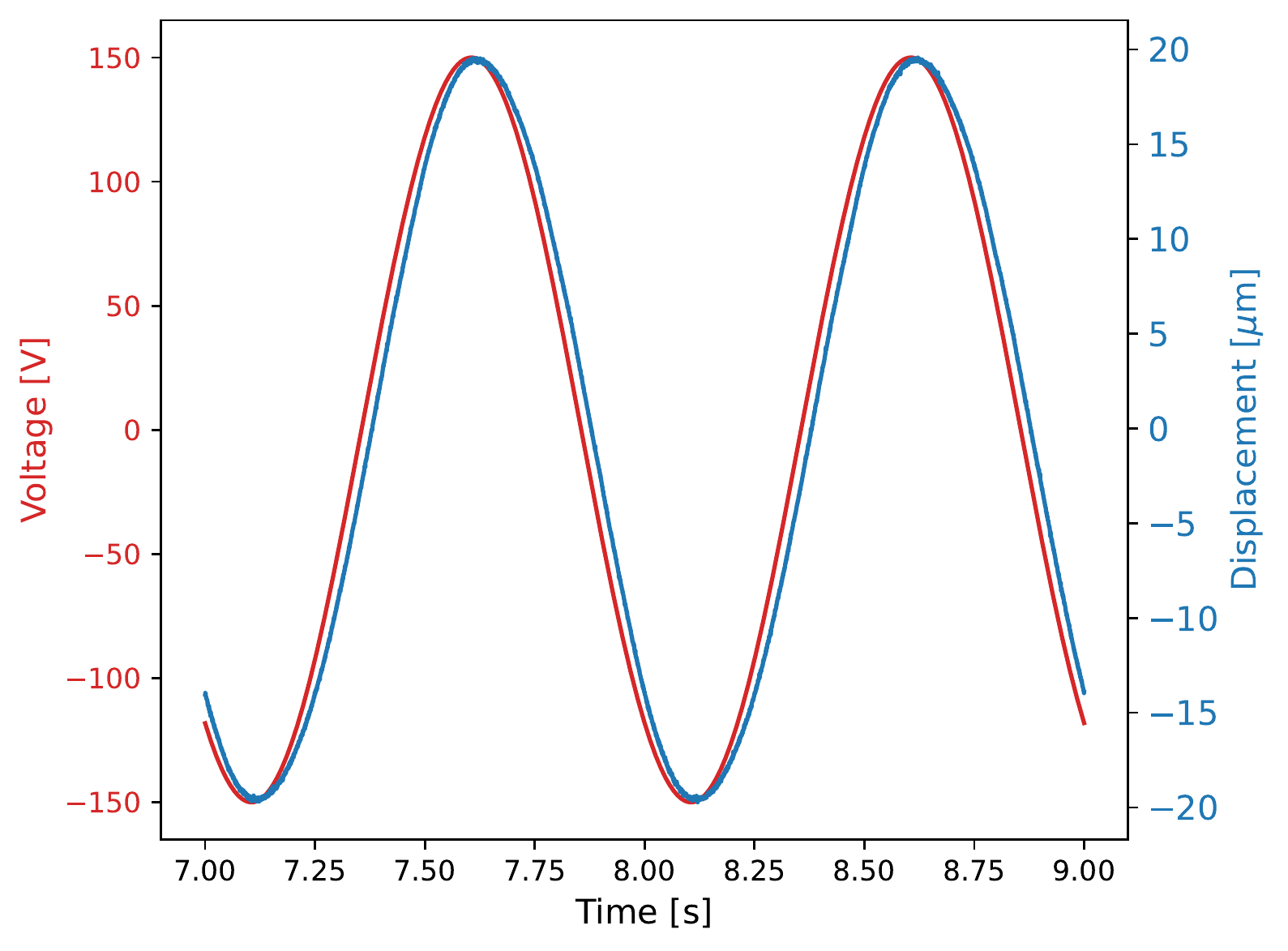}}

\subfloat{\includegraphics[clip,width=0.75\columnwidth]{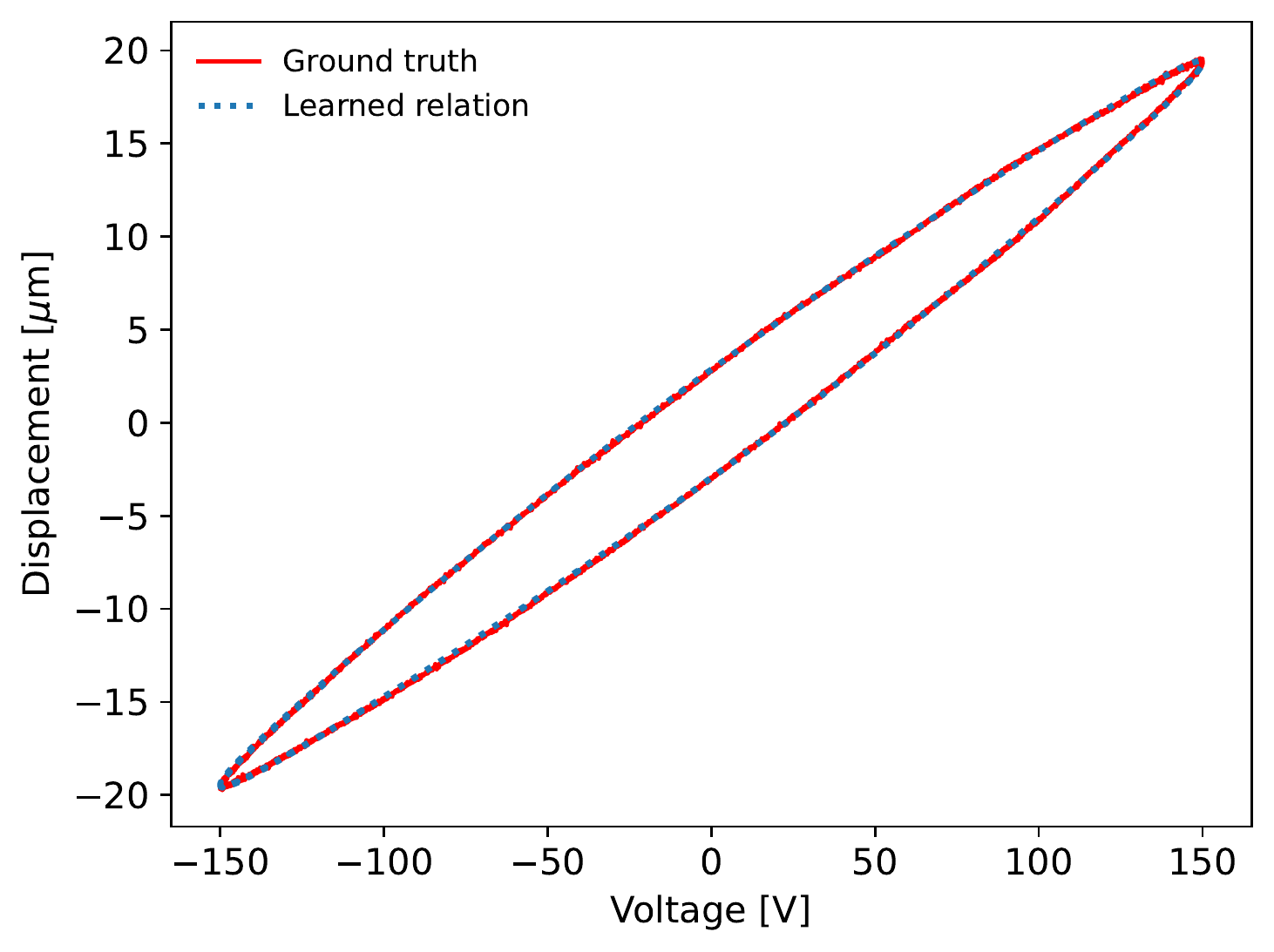}}

\subfloat{\includegraphics[clip,width=0.75\columnwidth]{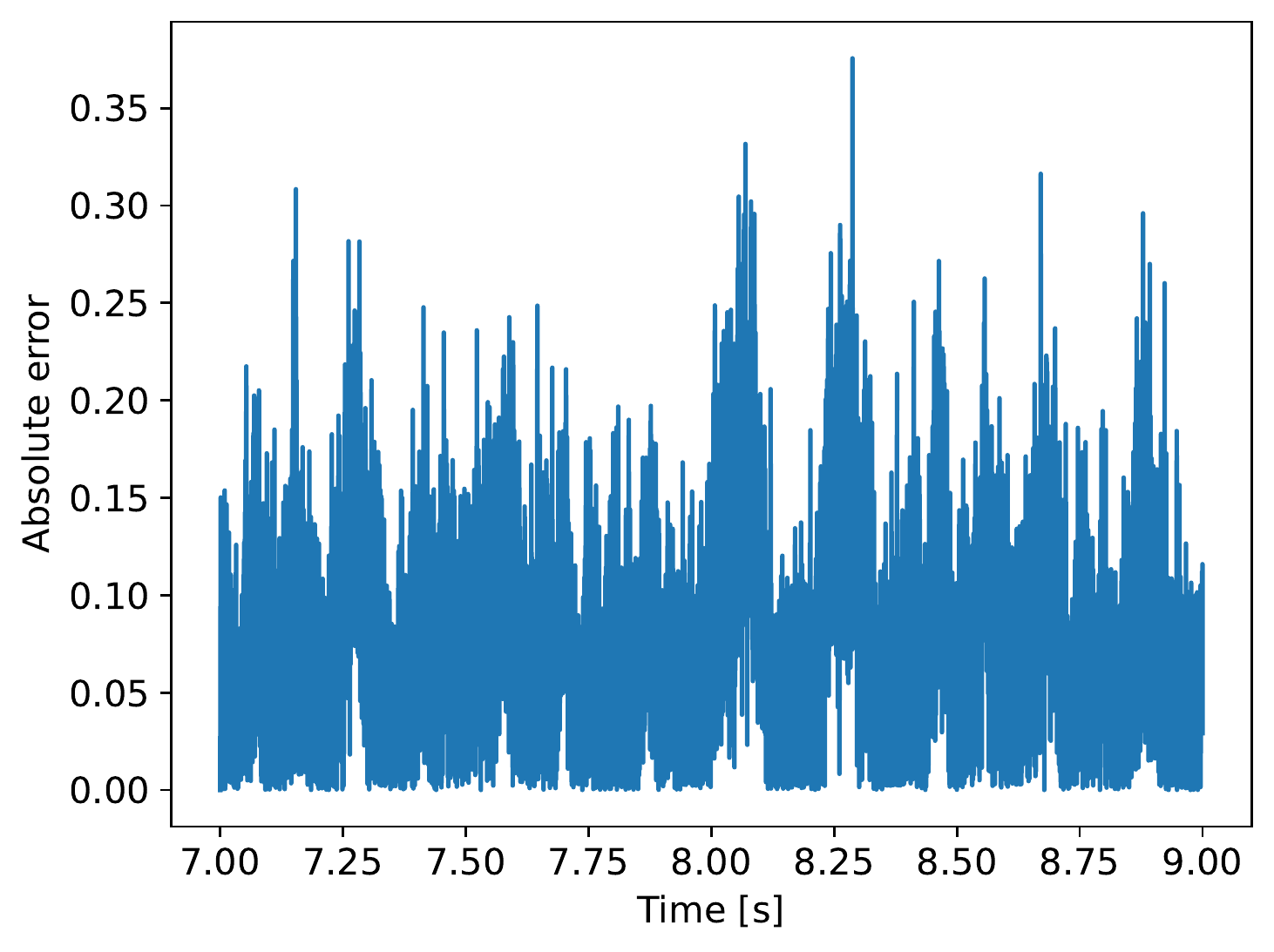}}
\caption{\textbf{Learning hysteresis model for real-world piezoelectric actuator \cite{ayala2020piezoelectric} (top:)} Input voltage and output displacement varying with respect to time. \textbf{(middle:)} Learned voltage vs displacement relationship compared against the experimental data acting as ground truth. \textbf{(bottom:)} Absolute error in the learned hysteresis model varying with respect to time.}
\label{fig3}
\end{figure}

However, measurement data may be corrupted in ways other than Gaussian noise. An alternative form of data corruption is the presence of outliers within the dataset, which are individual data points that exhibit substantial deviations from the standard pattern of the data. Outliers can emerge due to various factors, such as measurement errors, data corruption, or other forms of noise. When outliers are present in the dataset, they can distort the identification of the governing equations of the system, leading to inaccurate or unreliable results. SINDy works by minimizing a loss function that penalizes the complexity of the identified model, encouraging the selection of only the most important terms that capture the underlying dynamics. However, outliers can significantly increase the loss function and bias the selection of terms towards capturing the outliers, leading to overfitting and poor generalization. Therefore, it is important to preprocess the data and remove outliers before applying SINDy. Several outlier detection techniques can be used to identify and remove outliers from the data. Additionally, robust variants of SINDy \cite{tran2017exact, schaeffer2017sparse} that are designed to handle outliers can be used to improve the accuracy and reliability of the identified models.

\begin{table}
\caption{\label{tab:table1}Examining the performance of the proposed methodology to model hysteresis given a number of training points and comparing it with traditional regression methods and deep neural network using relative percent error as a metric.}
\begin{ruledtabular}
\begin{tabular}{cccc}
Method $\backslash$ Training size & 1000 & 5000 & 10000 \\
\hline
SINDy & $2.8 \times 10^{-4}$\% & $1  \times 10^{-5}$\% & $\sim$0\% \\ 
Linear Regression & $3.3 \times 10^{-4}$\% & $2.3  \times 10^{-5}$\% & $1 \times 10^{-3}$\%\\
Ridge Regression & $9 \times 10^{-2}$\% & $4.1 \times 10^{-3}$\% & $9 \times 10^{-3}$\% \\
Neural Network & $1.8 \times 10^{-3}$ \% & $4.4  \times 10^{-4}$\% & $3.9 \times 10^{-4}$\%  \\
\end{tabular}
\end{ruledtabular}
\end{table}

\begin{table}
\caption{\label{tab:table2}Examining the performance of the proposed methodology to model hysteresis for noisy training data in the form of Gaussian noise and comparing it with traditional regression methods and deep neural network using relative percent error as a metric.}
\begin{ruledtabular}
\begin{tabular}{ccc}
Method & Error for 1\% noise & Error for 5\% noise\\ 
\hline
SINDy & 0.0101\% & 0.25\%  \\
Linear Regression & 0.0104\% & 0.26\% \\
Ridge Regression & 1.46\% & 1.24\% \\
Neural Network & 1.54\% & 33.28\% \\
\end{tabular}
\end{ruledtabular}
\end{table}

The presented two comparisons on limited and noisy datasets assumed we know the feature library for linear regression, ridge regression and feedforward neural network in advance. However, in practice, this is not feasible. One could only assume the features to employ these methods in a real-world situation. For the next comparison, we take the output generated from the Duhem model but the features from the Bouc-Wen model, as the Bouc-Wen model is used more frequently to model hysteresis. \cite{ottosen2005mechanics} However, the library for SINDy remains the same. In such a scenario, SINDy outperforms the traditional methods exceptionally by achieving similar accuracy as the last case; however, the traditional methods approximate the dataset with more than a five percent error. Hence, SINDy proves suitable when no prior information about the model is known.

%\subsection{Piezoelectric actuator experimental dataset}
Next, we model a real-world piezoelectric actuator hysteresis dataset. An open-source voltage-displacement dataset \cite{ayala2020piezoelectric} was used to model a real-world dataset. The dataset is generated for a cantilever beam-shaped piezoelectric actuator used for positioning. The actuator was subjected to harmonic excitations, and $45000$ data points were available. The current study used only $15000$ data points as in \cite{ayala2020piezoelectric} without preprocessing with threshold $\lambda = 0.01$. However, several black-box treatments \cite{ayala2015nonlinear, ayala2020nonlinear} have been carried out for the dataset. This work presents the white-box dynamical system for the piezoelectric actuator given by,
\begin{equation}
    {\dot{w}(t) = -0.17 - 2.38 u + 0.58 \dot{u} + 0.12 |\dot{u}| - 0.07 w |w|}
\end{equation}
and the learned loop is presented in Fig. \ref{fig3} along with the experimental data. Fig. \ref{fig3} also presents the voltage-displacement data varying with time used for modelling. Also, the absolute error obtained in the learned model $|w-w^{*}|$ is presented against time in Fig. \ref{fig3}. The modelled equation represents the relationship between the displacement $w$ of a physical system and the voltage $u$ applied to it. Here, $\dot{w}$ represents the derivative of displacement with respect to time $t$, and $\dot{u}$ represents the derivative of voltage with respect to time. The physical interpretation of this equation can be broken down as follows:
\begin{itemize}
\item The first term, -$0.17$, represents a constant bias or offset that affects the rate of change of displacement, regardless of the voltage applied.
\item The second term, -$2.38u$, represents the effect of the voltage u on the rate of change of displacement. This term suggests that increasing the voltage will decrease the rate of change of displacement while decreasing the voltage will increase the rate of change of displacement.
\item The third term, $0.58\dot{u}$, represents the effect of the rate of change of voltage on the rate of change of displacement. This term suggests that increasing the rate of change of voltage will increase the rate of change of displacement.
\item The fourth term, $0.12|\dot{u}|$, represents the effect of the absolute value of the rate of change of voltage on the rate of change of displacement. This term suggests that an increase in the absolute value of the rate of change of voltage will increase the rate of change of displacement.
\item The fifth term, -$0.07w|w|$, represents the effect of the displacement on the rate of change of displacement. This term suggests that an increase in the absolute value of displacement will decrease the rate of change of displacement, while a decrease in the absolute value of displacement will increase the rate of change of displacement.
\end{itemize}

Overall, this equation describes how the displacement of the piezoelectric actuator responds to changes in the voltage applied to it and the voltage change rate. The specific coefficients in the equation determine the strength and direction of these relationships.

In addition, we compare our approximation results with the methods that have modelled this dataset in literature. An evolutionary algorithm \cite{ayala2015nonlinear} and a deep neural network-based method \cite{ayala2020nonlinear} are used to model the dataset with an $R2$ score of $0.99$. Our obtained $R2$ score is $0.99$ as well. However, it is well known that an evolutionary algorithm and DNN-based methods could take minutes and even hours to train, whereas our method takes seconds for training and testing. In another work, \cite{nguyen2021characterization} the authors modelled the dataset using four deep learning-based methods, namely, Long Short-Term Memory (LSTM), Gated Recurrent Unit (GRU), Temporal Convolutional Network (TCN) and Fully Convolutional Neural Network (FCN). Table \ref{tab:table3} presents the accuracy and computational time results. Normalized root mean square error (NRMSE) was chosen as the error metric as it was used in the previous study \cite{nguyen2021characterization} defined as

\begin{equation*}
\text{NRMSE} = \frac{\sqrt{\frac{1}{n}\sum_{i=1}^{n}(w_i - w^{*}_i)^2}}{w_{\max} - w_{\min}}
\end{equation*}
where $w^{*}$ is the simulated solution of the approximated model. $w_{\max}$ and $w_{\min}$ are the maximum and minimum values for the experimental data $w$. SINDy attained comparable accuracy to deep learning but could not beat TCN. This behavior is because of TCN's superior performance in modelling and predicting time series data that are memory dependent. The hysteresis data are itself memory dependent and hence is a suitable dataset for TCN. However, SINDy learned the model four orders of magnitude faster in terms of computational time. Computational time in modelling is important when the interest is in modelling and simulation simultaneously, where the overall computation time could be decreased using SINDy.

%The next section contains insights and recommendations to model hysteresis in magnetic materials through sparse regression techniques.

\begin{table}
\caption{\label{tab:table3}Performance of the proposed methodology to model hysteresis for the dataset of piezoelectric actuator presented by Ayala et al. \cite{ayala2020piezoelectric} and comparison with the results obtained by deep learning-based methods presented by Nguyen et al. \cite{nguyen2021characterization} using normalized root mean square error (NRMSE) as the metric.}
\begin{ruledtabular}
\begin{tabular}{cccc}
Method & NRMSE & Training Time & Testing Time\\
\hline
SINDy &  0.23 & $1.8 \times 10^{-2}$ s & $4.5 \times 10^{-4}$ s \\ 
LSTM \cite{nguyen2021characterization} &  0.25 & $104.83$ s & $7.1 \times 10^{-3}$ s\\
GRU \cite{nguyen2021characterization} &  0.3 & $101.31$ s & $5.4 \times 10^{-3}$ s \\
TCN \cite{nguyen2021characterization} &  0.19 & $61.57$ s & $3.0 \times 10^{-3}$ s\\ 
FCN \cite{nguyen2021characterization} &  0.6 & $74.92$ s & $3.4 \times 10^{-3}$ s\\
\end{tabular}
\end{ruledtabular}
\end{table}

The presented experiments so far dealt with single-loop hysteresis. However, hysteresis loops fall into two categories: single and multiple. Single-loop hysteresis is common in literature for piezoelectric materials, while multiple-loop hysteresis exhibits complex and nonlinear behaviour, with two branches resembling butterfly wings. The butterfly hysteresis loop comprises two single hysteresis loops with both clockwise and counter-clockwise directions. However, constructing a suitable model to explain the butterfly hysteresis effect is challenging due to the lack of available hysteresis models for this type of behaviour. Unlike single-loop hysteresis, which only exhibits one direction, the butterfly hysteresis effect has both directions, making it challenging to create mathematical models that can accurately describe it. Following, we aim to model butterfly-shaped hysteresis through the proposed methodology.

\begin{figure}[htp]
\subfloat{\includegraphics[clip,width=0.75\columnwidth]{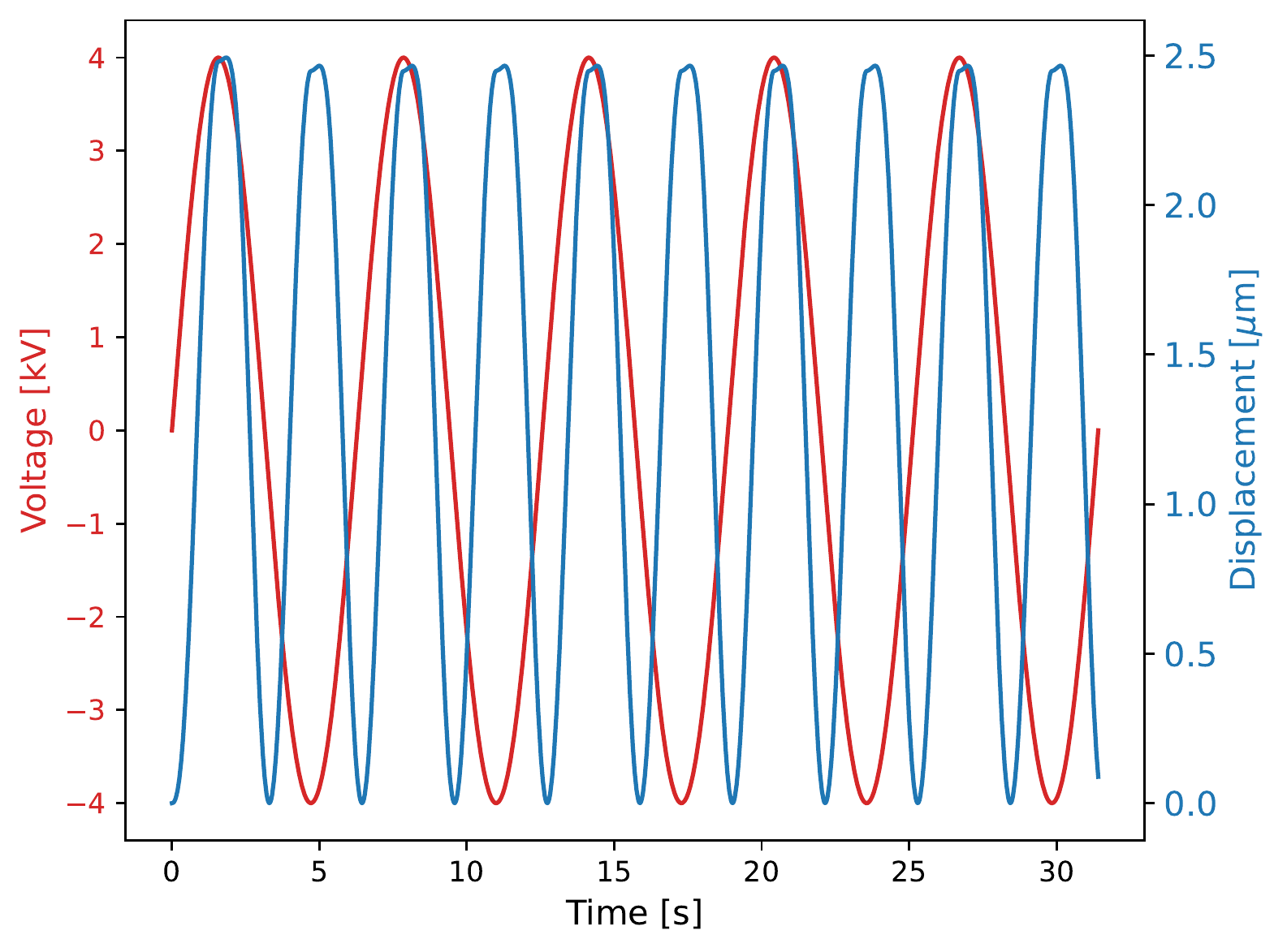}}

\subfloat{\includegraphics[clip,width=0.75\columnwidth]{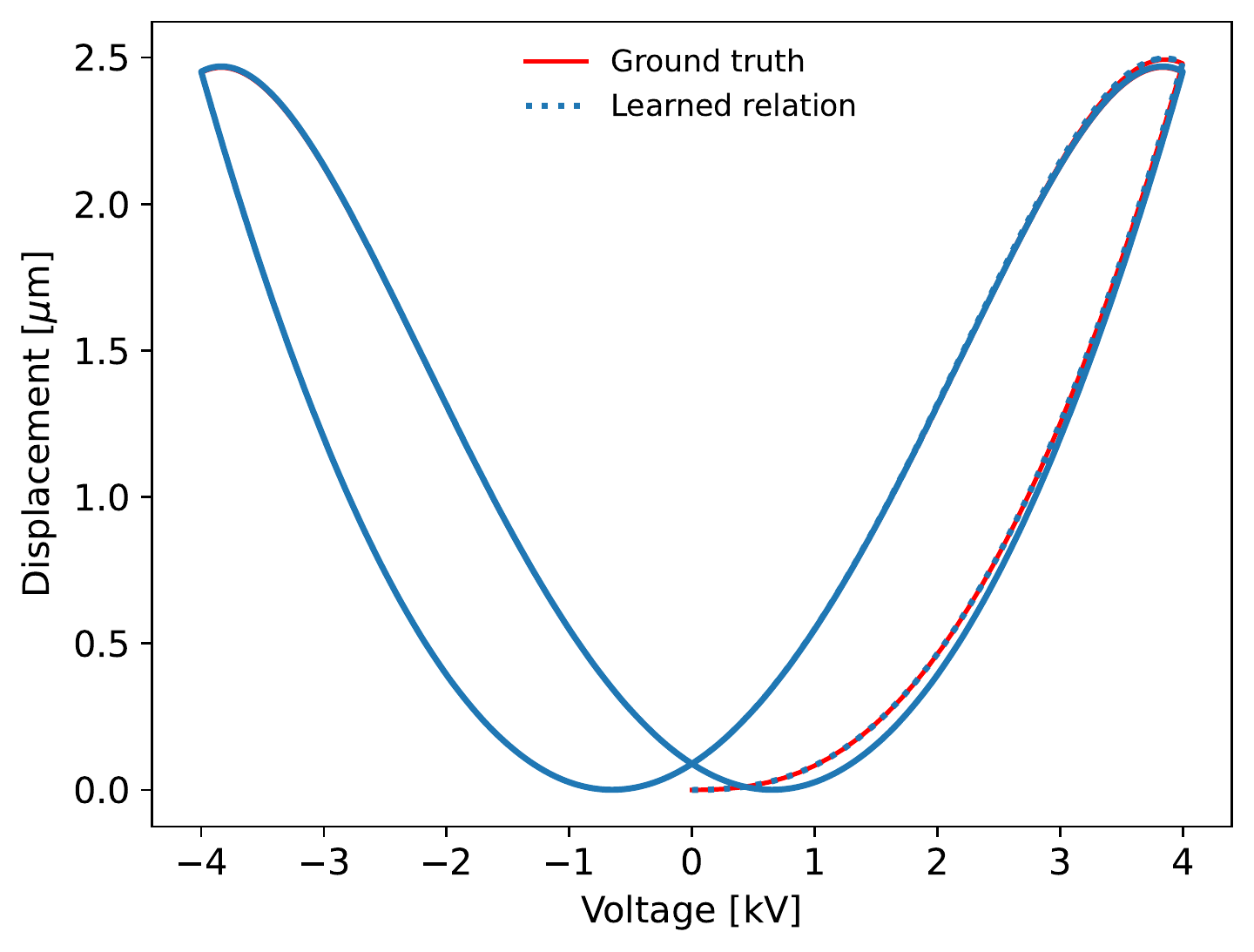}}
\caption{\textbf{Learning butterfly-shaped hysteresis model from the simulated data (top:)} Input voltage and output displacement varying with respect to time. \textbf{(bottom:)} Learned voltage vs displacement relationship compared against the simulated data acting as ground truth.}
\label{fig4}
\end{figure}

The dataset exhibits a butterfly-shaped pattern, learned using a threshold of $\lambda=0.1$. Fig. \ref{fig4} displays the input voltage and output displacement variation over time and compares the learned voltage-displacement relationship and the data employed for modelling. The learned butterfly hysteresis model is given by

\begin{equation}
\begin{split}
\dot{w}(t) = 6.4 |\dot{u}(t)| u(t) y(t) - 3.4 |\dot{u}(t)| w(t) y(t) \\ + 0.8 \dot{u}(t) y(t)
\end{split}
\end{equation}
where $w(t)$ is the output displacement and $u(t)$ is the input voltage. $y(t)$ is the output obtained from the Bouc-Wen model 

\begin{equation}
     \dot{y}(t) = 3.2 |\dot{u}(t)| u(t) - 1.7 |\dot{u}(t)| y(t) + 0.4 \dot{u}(t)
\end{equation}

The hysteresis is well-learned with the proposed methodology, as indicated by $\mathcal{R} = 3.1 \times 10^{-4}$ for the butterfly-shaped hysteresis. One could infer that $\dot{w} = 2y\dot{y}$ by examining the model obtained for the butterfly-shaped loop. Hence, a butterfly-shaped loop is a generalization of the Duhem model itself and could be found through an intermediate step of the first modelling single hysteresis loop. Also, this suggests that a general butterfly-shaped hysteresis loop could be expressed as $w = y^{2} + c$ where $c$ is a constant. In literature, complicated analytical relationships have been derived to model this relationship.\cite{li2021development} However, this physical insight could potentially help tackle more complicated butterfly responses in the future. Also, as the considered Duhem model is known to model symmetrical loops, similar generalizations of asymmetric Duhem type models could help model asymmetric butterfly-shaped hysteresis loops.

%\section{Discussion on modelling magnetic hysteresis}
% Next, an experiment on non-oriented electric steel (NO27) was carried out to test the validity of the proposed method for magnetic materials. The hysteresis loop measured in this work was obtained by an Epstein frame, calibrated to comply with the IEC standard, \cite{ieee} of which the core was constructed out of $16$ strips of NO27-1450H. Only AC measurements were performed, for which the secondary voltage, i.e. the magnetic flux density waveform, was controlled to remain sinusoidal. The measurements were conducted for a magnetic flux density of $0.2$ T at a frequency of $10$ Hz. The modelled differential equation with threshold $\lambda = 0.01$ is,
% \begin{eqnarray*}
% \dot{B} = -0.814 B + 0.651 H + 2.983 \dot{H} + 0.063 B |B|
%  - 0.026 H |\dot{H}| \\ - 0.058 H |B| - 0.132 \dot{H} |\dot{H}| - 0.061 |\dot{H}| |B| 
% \end{eqnarray*}

Finally, we present the conclusions and insights obtained from the study. The following conclusions could be drawn from this study. This paper introduced an approach for modelling hysteresis in piezoelectric materials using a sparsity-promoting machine learning technique. The STLSQ algorithm was applied to assist in sparsity during the learning process. The algorithm only requires input-output data and accurately discovers the white-box hysteresis model with minimal error. The relationship was expressed as a dynamic system, and the resulting ordinary differential equation (ODE) could be simulated using any standard ODE solver to predict hysteresis. Additionally, the method did not require prior assumptions or knowledge about the underlying model governing hysteresis during the modelling phase.

The presented algorithm was tested on a small training set with just $1000$ data points with an error of less than $1 \times 10^{-3}$ percent. Furthermore, the method was validated on raw experimental and noisy simulated data with up to $5$ percent Gaussian noise. The algorithm proved accurate for an open-source piezoelectric actuator dataset. It achieved the fastest computational time compared to state-of-the-art methods for the same dataset in the literature. The presented methodology also modelled the butterfly-shaped hysteresis efficiently. Overall, the presented numerical experiments indicate the sparse discovery of hysteresis models for piezoelectric materials to be accurate, robust and computationally inexpensive.

%\begin{acknowledgments}
%We wish to acknowledge the support of the author community in using
%REV\TeX{}, offering suggestions and encouragement, testing new versions,
%\end{acknowledgments}
\section*{Author Declarations}

\subsection*{Conflict of Interest}

The authors have no conflicts to disclose.

\subsection*{Author Contributions}

{\textbf{Abhishek Chandra:}} Conceptualization; Data curation; Formal analysis; Investigation; Validation; Software; Writing – original draft; Writing – review $\&$ editing. {\textbf{Bram Daniels:}} Writing – review $\&$ editing. {\textbf{Mitrofan Curti:}} Supervision; Funding acquisition; Writing – review $\&$ editing. {\textbf{Koen Tiels:}} Supervision; Funding acquisition; Writing – review $\&$ editing. {\textbf{Elena A. Lomonova:}} Resources; Supervision; Funding acquisition; Writing – review $\&$ editing. {\textbf{Daniel M. Tartakovsky:}} Supervision; Writing – review $\&$ editing.

\subsection*{Data Availability Statement}

The data that support the findings of this study are openly available at \href{https://github.com/chandratue/SmartHysteresis}{https://github.com/chandratue/SmartHysteresis.}

\nocite{*}
\bibliography{aipsamp}% Produces the bibliography via BibTeX.

\providecommand{\noopsort}[1]{}\providecommand{\singleletter}[1]{#1}%
\begin{thebibliography}{23}%
\makeatletter
\providecommand \@ifxundefined [1]{%
 \@ifx{#1\undefined}
}%
\providecommand \@ifnum [1]{%
 \ifnum #1\expandafter \@firstoftwo
 \else \expandafter \@secondoftwo
 \fi
}%
\providecommand \@ifx [1]{%
 \ifx #1\expandafter \@firstoftwo
 \else \expandafter \@secondoftwo
 \fi
}%
\providecommand \natexlab [1]{#1}%
\providecommand \enquote  [1]{``#1''}%
\providecommand \bibnamefont  [1]{#1}%
\providecommand \bibfnamefont [1]{#1}%
\providecommand \citenamefont [1]{#1}%
\providecommand \href@noop [0]{\@secondoftwo}%
\providecommand \href [0]{\begingroup \@sanitize@url \@href}%
\providecommand \@href[1]{\@@startlink{#1}\@@href}%
\providecommand \@@href[1]{\endgroup#1\@@endlink}%
\providecommand \@sanitize@url [0]{\catcode `\\12\catcode `\$12\catcode
  `\&12\catcode `\#12\catcode `\^12\catcode `\_12\catcode `\%12\relax}%
\providecommand \@@startlink[1]{}%
\providecommand \@@endlink[0]{}%
\providecommand \url  [0]{\begingroup\@sanitize@url \@url }%
\providecommand \@url [1]{\endgroup\@href {#1}{\urlprefix }}%
\providecommand \urlprefix  [0]{URL }%
\providecommand \Eprint [0]{\href }%
\providecommand \doibase [0]{http://dx.doi.org/}%
\providecommand \selectlanguage [0]{\@gobble}%
\providecommand \bibinfo  [0]{\@secondoftwo}%
\providecommand \bibfield  [0]{\@secondoftwo}%
\providecommand \translation [1]{[#1]}%
\providecommand \BibitemOpen [0]{}%
\providecommand \bibitemStop [0]{}%
\providecommand \bibitemNoStop [0]{.\EOS\space}%
\providecommand \EOS [0]{\spacefactor3000\relax}%
\providecommand \BibitemShut  [1]{\csname bibitem#1\endcsname}%
\let\auto@bib@innerbib\@empty
%</preamble>
\bibitem [{\citenamefont {Ottosen}\ and\ \citenamefont
  {Ristinmaa}(2005)}]{ottosen2005mechanics}%
  \BibitemOpen
  \bibfield  {author} {\bibinfo {author} {\bibfnamefont {N.~S.}\ \bibnamefont
  {Ottosen}}\ and\ \bibinfo {author} {\bibfnamefont {M.}~\bibnamefont
  {Ristinmaa}},\ }\href@noop {} {\emph {\bibinfo {title} {The mechanics of
  constitutive modeling}}}\ (\bibinfo  {publisher} {Elsevier},\ \bibinfo {year}
  {2005})\BibitemShut {NoStop}%
\bibitem [{\citenamefont {Bertotti}(1998)}]{bertotti1998hysteresis}%
  \BibitemOpen
  \bibfield  {author} {\bibinfo {author} {\bibfnamefont {G.}~\bibnamefont
  {Bertotti}},\ }\href@noop {} {\emph {\bibinfo {title} {Hysteresis in
  magnetism: for physicists, materials scientists, and engineers}}}\ (\bibinfo
  {publisher} {Gulf Professional Publishing},\ \bibinfo {year}
  {1998})\BibitemShut {NoStop}%
\bibitem [{\citenamefont {Lin}\ and\ \citenamefont
  {Lin}(2012)}]{lin2012tracking}%
  \BibitemOpen
  \bibfield  {author} {\bibinfo {author} {\bibfnamefont {C.-J.}\ \bibnamefont
  {Lin}}\ and\ \bibinfo {author} {\bibfnamefont {P.-T.}\ \bibnamefont {Lin}},\
  }\bibfield  {title} {\enquote {\bibinfo {title} {Tracking control of a
  biaxial piezo-actuated positioning stage using generalized duhem model},}\
  }\href@noop {} {\bibfield  {journal} {\bibinfo  {journal} {Computers \&
  Mathematics with Applications}\ }\textbf {\bibinfo {volume} {64}},\ \bibinfo
  {pages} {766--787} (\bibinfo {year} {2012})}\BibitemShut {NoStop}%
\bibitem [{\citenamefont {Gan}\ and\ \citenamefont
  {Zhang}(2019)}]{gan2019review}%
  \BibitemOpen
  \bibfield  {author} {\bibinfo {author} {\bibfnamefont {J.}~\bibnamefont
  {Gan}}\ and\ \bibinfo {author} {\bibfnamefont {X.}~\bibnamefont {Zhang}},\
  }\bibfield  {title} {\enquote {\bibinfo {title} {A review of nonlinear
  hysteresis modeling and control of piezoelectric actuators},}\ }\href@noop {}
  {\bibfield  {journal} {\bibinfo  {journal} {AIP Advances}\ }\textbf {\bibinfo
  {volume} {9}},\ \bibinfo {pages} {040702} (\bibinfo {year}
  {2019})}\BibitemShut {NoStop}%
\bibitem [{\citenamefont {Bertotti}\ and\ \citenamefont
  {Mayergoyz}(2005)}]{bertotti2005science}%
  \BibitemOpen
  \bibfield  {author} {\bibinfo {author} {\bibfnamefont {G.}~\bibnamefont
  {Bertotti}}\ and\ \bibinfo {author} {\bibfnamefont {I.~D.}\ \bibnamefont
  {Mayergoyz}},\ }\href@noop {} {\emph {\bibinfo {title} {The Science of
  Hysteresis: 3-volume set}}}\ (\bibinfo  {publisher} {Elsevier},\ \bibinfo
  {year} {2005})\BibitemShut {NoStop}%
\bibitem [{\citenamefont {Zeinali}, \citenamefont {Krop},\ and\ \citenamefont
  {Lomonova}(2019)}]{zeinali2019comparison}%
  \BibitemOpen
  \bibfield  {author} {\bibinfo {author} {\bibfnamefont {R.}~\bibnamefont
  {Zeinali}}, \bibinfo {author} {\bibfnamefont {D.~C.}\ \bibnamefont {Krop}}, \
  and\ \bibinfo {author} {\bibfnamefont {E.~A.}\ \bibnamefont {Lomonova}},\
  }\bibfield  {title} {\enquote {\bibinfo {title} {Comparison of preisach and
  congruency-based static hysteresis models applied to non-oriented steels},}\
  }\href@noop {} {\bibfield  {journal} {\bibinfo  {journal} {IEEE Transactions
  on Magnetics}\ }\textbf {\bibinfo {volume} {56}},\ \bibinfo {pages} {1--4}
  (\bibinfo {year} {2019})}\BibitemShut {NoStop}%
\bibitem [{\citenamefont {Daniels}\ \emph {et~al.}(2023)\citenamefont
  {Daniels}, \citenamefont {Overboom}, \citenamefont {Curti},\ and\
  \citenamefont {Lomonova}}]{10034669}%
  \BibitemOpen
  \bibfield  {author} {\bibinfo {author} {\bibfnamefont {B.}~\bibnamefont
  {Daniels}}, \bibinfo {author} {\bibfnamefont {T.}~\bibnamefont {Overboom}},
  \bibinfo {author} {\bibfnamefont {M.}~\bibnamefont {Curti}}, \ and\ \bibinfo
  {author} {\bibfnamefont {E.}~\bibnamefont {Lomonova}},\ }\bibfield  {title}
  {\enquote {\bibinfo {title} {Everett map construction for modeling static
  hysteresis: Delaunay based interpolant versus b-spline surface},}\ }\href
  {\doibase 10.1109/TMAG.2023.3241427} {\bibfield  {journal} {\bibinfo
  {journal} {IEEE Transactions on Magnetics}\ ,\ \bibinfo {pages} {1--1}}
  (\bibinfo {year} {2023})}\BibitemShut {NoStop}%
\bibitem [{\citenamefont {Kim}, \citenamefont {Kwon},\ and\ \citenamefont
  {Song}(2019)}]{kim2019response}%
  \BibitemOpen
  \bibfield  {author} {\bibinfo {author} {\bibfnamefont {T.}~\bibnamefont
  {Kim}}, \bibinfo {author} {\bibfnamefont {O.-S.}\ \bibnamefont {Kwon}}, \
  and\ \bibinfo {author} {\bibfnamefont {J.}~\bibnamefont {Song}},\ }\bibfield
  {title} {\enquote {\bibinfo {title} {Response prediction of nonlinear
  hysteretic systems by deep neural networks},}\ }\href@noop {} {\bibfield
  {journal} {\bibinfo  {journal} {Neural Networks}\ }\textbf {\bibinfo {volume}
  {111}},\ \bibinfo {pages} {1--10} (\bibinfo {year} {2019})}\BibitemShut
  {NoStop}%
\bibitem [{\citenamefont {Wei}\ and\ \citenamefont
  {Sun}(2000)}]{wei2000constructing}%
  \BibitemOpen
  \bibfield  {author} {\bibinfo {author} {\bibfnamefont {J.-D.}\ \bibnamefont
  {Wei}}\ and\ \bibinfo {author} {\bibfnamefont {C.-T.}\ \bibnamefont {Sun}},\
  }\bibfield  {title} {\enquote {\bibinfo {title} {Constructing hysteretic
  memory in neural networks},}\ }\href@noop {} {\bibfield  {journal} {\bibinfo
  {journal} {IEEE Transactions on Systems, Man, and Cybernetics, Part B
  (Cybernetics)}\ }\textbf {\bibinfo {volume} {30}},\ \bibinfo {pages}
  {601--609} (\bibinfo {year} {2000})}\BibitemShut {NoStop}%
\bibitem [{\citenamefont {Mont{\'a}ns}\ \emph {et~al.}(2019)\citenamefont
  {Mont{\'a}ns}, \citenamefont {Chinesta}, \citenamefont
  {G{\'o}mez-Bombarelli},\ and\ \citenamefont {Kutz}}]{montans2019data}%
  \BibitemOpen
  \bibfield  {author} {\bibinfo {author} {\bibfnamefont {F.~J.}\ \bibnamefont
  {Mont{\'a}ns}}, \bibinfo {author} {\bibfnamefont {F.}~\bibnamefont
  {Chinesta}}, \bibinfo {author} {\bibfnamefont {R.}~\bibnamefont
  {G{\'o}mez-Bombarelli}}, \ and\ \bibinfo {author} {\bibfnamefont {J.~N.}\
  \bibnamefont {Kutz}},\ }\bibfield  {title} {\enquote {\bibinfo {title}
  {Data-driven modeling and learning in science and engineering},}\ }\href@noop
  {} {\bibfield  {journal} {\bibinfo  {journal} {Comptes Rendus M{\'e}canique}\
  }\textbf {\bibinfo {volume} {347}},\ \bibinfo {pages} {845--855} (\bibinfo
  {year} {2019})}\BibitemShut {NoStop}%
\bibitem [{\citenamefont {Kyprianou}, \citenamefont {Worden},\ and\
  \citenamefont {Panet}(2001)}]{kyprianou2001identification}%
  \BibitemOpen
  \bibfield  {author} {\bibinfo {author} {\bibfnamefont {A.}~\bibnamefont
  {Kyprianou}}, \bibinfo {author} {\bibfnamefont {K.}~\bibnamefont {Worden}}, \
  and\ \bibinfo {author} {\bibfnamefont {M.}~\bibnamefont {Panet}},\ }\bibfield
   {title} {\enquote {\bibinfo {title} {Identification of hysteretic systems
  using the differential evolution algorithm},}\ }\href@noop {} {\bibfield
  {journal} {\bibinfo  {journal} {Journal of Sound and vibration}\ }\textbf
  {\bibinfo {volume} {248}},\ \bibinfo {pages} {289--314} (\bibinfo {year}
  {2001})}\BibitemShut {NoStop}%
\bibitem [{\citenamefont {Chandra}\ \emph {et~al.}(2022)\citenamefont
  {Chandra}, \citenamefont {Curti}, \citenamefont {Tiels}, \citenamefont
  {Lomonova},\ and\ \citenamefont {Tartakovsky}}]{chandra2022data}%
  \BibitemOpen
  \bibfield  {author} {\bibinfo {author} {\bibfnamefont {A.}~\bibnamefont
  {Chandra}}, \bibinfo {author} {\bibfnamefont {M.}~\bibnamefont {Curti}},
  \bibinfo {author} {\bibfnamefont {K.}~\bibnamefont {Tiels}}, \bibinfo
  {author} {\bibfnamefont {E.~A.}\ \bibnamefont {Lomonova}}, \ and\ \bibinfo
  {author} {\bibfnamefont {D.~M.}\ \bibnamefont {Tartakovsky}},\ }\bibfield
  {title} {\enquote {\bibinfo {title} {Data-driven sparse discovery of
  hysteresis models for piezoelectric actuators},}\ }in\ \href@noop {} {\emph
  {\bibinfo {booktitle} {2022 IEEE 20th Biennial Conference on Electromagnetic
  Field Computation (CEFC)}}}\ (\bibinfo {organization} {IEEE},\ \bibinfo
  {year} {2022})\ pp.\ \bibinfo {pages} {1--2}\BibitemShut {NoStop}%
\bibitem [{\citenamefont {Brunton}, \citenamefont {Proctor},\ and\
  \citenamefont {Kutz}(2016)}]{brunton2016discovering}%
  \BibitemOpen
  \bibfield  {author} {\bibinfo {author} {\bibfnamefont {S.~L.}\ \bibnamefont
  {Brunton}}, \bibinfo {author} {\bibfnamefont {J.~L.}\ \bibnamefont
  {Proctor}}, \ and\ \bibinfo {author} {\bibfnamefont {J.~N.}\ \bibnamefont
  {Kutz}},\ }\bibfield  {title} {\enquote {\bibinfo {title} {Discovering
  governing equations from data by sparse identification of nonlinear dynamical
  systems},}\ }\href@noop {} {\bibfield  {journal} {\bibinfo  {journal}
  {Proceedings of the national academy of sciences}\ }\textbf {\bibinfo
  {volume} {113}},\ \bibinfo {pages} {3932--3937} (\bibinfo {year}
  {2016})}\BibitemShut {NoStop}%
\bibitem [{\citenamefont {Lai}\ and\ \citenamefont
  {Nagarajaiah}(2019)}]{lai2019sparse}%
  \BibitemOpen
  \bibfield  {author} {\bibinfo {author} {\bibfnamefont {Z.}~\bibnamefont
  {Lai}}\ and\ \bibinfo {author} {\bibfnamefont {S.}~\bibnamefont
  {Nagarajaiah}},\ }\bibfield  {title} {\enquote {\bibinfo {title} {Sparse
  structural system identification method for nonlinear dynamic systems with
  hysteresis/inelastic behavior},}\ }\href@noop {} {\bibfield  {journal}
  {\bibinfo  {journal} {Mechanical Systems and Signal Processing}\ }\textbf
  {\bibinfo {volume} {117}},\ \bibinfo {pages} {813--842} (\bibinfo {year}
  {2019})}\BibitemShut {NoStop}%
\bibitem [{\citenamefont {Brunton}\ and\ \citenamefont
  {Kutz}(2019)}]{brunton2019data}%
  \BibitemOpen
  \bibfield  {author} {\bibinfo {author} {\bibfnamefont {S.~L.}\ \bibnamefont
  {Brunton}}\ and\ \bibinfo {author} {\bibfnamefont {J.~N.}\ \bibnamefont
  {Kutz}},\ }\href@noop {} {\emph {\bibinfo {title} {Data-driven science and
  engineering: Machine learning, dynamical systems, and control}}}\ (\bibinfo
  {publisher} {Cambridge University Press},\ \bibinfo {year}
  {2019})\BibitemShut {NoStop}%
\bibitem [{\citenamefont {de~Silva}\ \emph {et~al.}(2020)\citenamefont
  {de~Silva}, \citenamefont {Champion}, \citenamefont {Quade}, \citenamefont
  {Loiseau}, \citenamefont {Kutz},\ and\ \citenamefont
  {Brunton}}]{de2020pysindy}%
  \BibitemOpen
  \bibfield  {author} {\bibinfo {author} {\bibfnamefont {B.~M.}\ \bibnamefont
  {de~Silva}}, \bibinfo {author} {\bibfnamefont {K.}~\bibnamefont {Champion}},
  \bibinfo {author} {\bibfnamefont {M.}~\bibnamefont {Quade}}, \bibinfo
  {author} {\bibfnamefont {J.-C.}\ \bibnamefont {Loiseau}}, \bibinfo {author}
  {\bibfnamefont {J.~N.}\ \bibnamefont {Kutz}}, \ and\ \bibinfo {author}
  {\bibfnamefont {S.~L.}\ \bibnamefont {Brunton}},\ }\bibfield  {title}
  {\enquote {\bibinfo {title} {{PySINDy}: a {Python} package for the sparse
  identification of nonlinear dynamics from data},}\ }\href@noop {} {\bibfield
  {journal} {\bibinfo  {journal} {arXiv preprint arXiv:2004.08424}\ } (\bibinfo
  {year} {2020})}\BibitemShut {NoStop}%
\bibitem [{\citenamefont {Ayala}, \citenamefont {Rakotondrabe},\ and\
  \citenamefont {dos Santos~Coelho}(2020)}]{ayala2020piezoelectric}%
  \BibitemOpen
  \bibfield  {author} {\bibinfo {author} {\bibfnamefont {H.~V.~H.}\
  \bibnamefont {Ayala}}, \bibinfo {author} {\bibfnamefont {M.}~\bibnamefont
  {Rakotondrabe}}, \ and\ \bibinfo {author} {\bibfnamefont {L.}~\bibnamefont
  {dos Santos~Coelho}},\ }\bibfield  {title} {\enquote {\bibinfo {title}
  {Piezoelectric micromanipulator dataset for hysteresis identification},}\
  }\href@noop {} {\bibfield  {journal} {\bibinfo  {journal} {Data in brief}\
  }\textbf {\bibinfo {volume} {29}},\ \bibinfo {pages} {105175} (\bibinfo
  {year} {2020})}\BibitemShut {NoStop}%
\bibitem [{\citenamefont {Tran}\ and\ \citenamefont
  {Ward}(2017)}]{tran2017exact}%
  \BibitemOpen
  \bibfield  {author} {\bibinfo {author} {\bibfnamefont {G.}~\bibnamefont
  {Tran}}\ and\ \bibinfo {author} {\bibfnamefont {R.}~\bibnamefont {Ward}},\
  }\bibfield  {title} {\enquote {\bibinfo {title} {Exact recovery of chaotic
  systems from highly corrupted data},}\ }\href@noop {} {\bibfield  {journal}
  {\bibinfo  {journal} {Multiscale Modeling \& Simulation}\ }\textbf {\bibinfo
  {volume} {15}},\ \bibinfo {pages} {1108--1129} (\bibinfo {year}
  {2017})}\BibitemShut {NoStop}%
\bibitem [{\citenamefont {Schaeffer}\ and\ \citenamefont
  {McCalla}(2017)}]{schaeffer2017sparse}%
  \BibitemOpen
  \bibfield  {author} {\bibinfo {author} {\bibfnamefont {H.}~\bibnamefont
  {Schaeffer}}\ and\ \bibinfo {author} {\bibfnamefont {S.~G.}\ \bibnamefont
  {McCalla}},\ }\bibfield  {title} {\enquote {\bibinfo {title} {Sparse model
  selection via integral terms},}\ }\href@noop {} {\bibfield  {journal}
  {\bibinfo  {journal} {Physical Review E}\ }\textbf {\bibinfo {volume} {96}},\
  \bibinfo {pages} {023302} (\bibinfo {year} {2017})}\BibitemShut {NoStop}%
\bibitem [{\citenamefont {Ayala}\ \emph {et~al.}(2015)\citenamefont {Ayala},
  \citenamefont {Habineza}, \citenamefont {Rakotondrabe}, \citenamefont
  {Klein},\ and\ \citenamefont {Coelho}}]{ayala2015nonlinear}%
  \BibitemOpen
  \bibfield  {author} {\bibinfo {author} {\bibfnamefont {H.~V.~H.}\
  \bibnamefont {Ayala}}, \bibinfo {author} {\bibfnamefont {D.}~\bibnamefont
  {Habineza}}, \bibinfo {author} {\bibfnamefont {M.}~\bibnamefont
  {Rakotondrabe}}, \bibinfo {author} {\bibfnamefont {C.~E.}\ \bibnamefont
  {Klein}}, \ and\ \bibinfo {author} {\bibfnamefont {L.~S.}\ \bibnamefont
  {Coelho}},\ }\bibfield  {title} {\enquote {\bibinfo {title} {Nonlinear
  black-box system identification through neural networks of a hysteretic
  piezoelectric robotic micromanipulator},}\ }\href@noop {} {\bibfield
  {journal} {\bibinfo  {journal} {IFAC-PapersOnLine}\ }\textbf {\bibinfo
  {volume} {48}},\ \bibinfo {pages} {409--414} (\bibinfo {year}
  {2015})}\BibitemShut {NoStop}%
\bibitem [{\citenamefont {Ayala}\ \emph {et~al.}(2020)\citenamefont {Ayala},
  \citenamefont {Habineza}, \citenamefont {Rakotondrabe},\ and\ \citenamefont
  {dos Santos~Coelho}}]{ayala2020nonlinear}%
  \BibitemOpen
  \bibfield  {author} {\bibinfo {author} {\bibfnamefont {H.~V.~H.}\
  \bibnamefont {Ayala}}, \bibinfo {author} {\bibfnamefont {D.}~\bibnamefont
  {Habineza}}, \bibinfo {author} {\bibfnamefont {M.}~\bibnamefont
  {Rakotondrabe}}, \ and\ \bibinfo {author} {\bibfnamefont {L.}~\bibnamefont
  {dos Santos~Coelho}},\ }\bibfield  {title} {\enquote {\bibinfo {title}
  {Nonlinear black-box system identification through coevolutionary algorithms
  and radial basis function artificial neural networks},}\ }\href@noop {}
  {\bibfield  {journal} {\bibinfo  {journal} {Applied Soft Computing}\ }\textbf
  {\bibinfo {volume} {87}},\ \bibinfo {pages} {105990} (\bibinfo {year}
  {2020})}\BibitemShut {NoStop}%
\bibitem [{\citenamefont {Nguyen}\ and\ \citenamefont
  {Chauhan}(2021)}]{nguyen2021characterization}%
  \BibitemOpen
  \bibfield  {author} {\bibinfo {author} {\bibfnamefont {X.~A.}\ \bibnamefont
  {Nguyen}}\ and\ \bibinfo {author} {\bibfnamefont {S.}~\bibnamefont
  {Chauhan}},\ }\bibfield  {title} {\enquote {\bibinfo {title}
  {Characterization of flexible and stretchable sensors using neural
  networks},}\ }\href@noop {} {\bibfield  {journal} {\bibinfo  {journal}
  {Measurement Science and Technology}\ }\textbf {\bibinfo {volume} {32}},\
  \bibinfo {pages} {075004} (\bibinfo {year} {2021})}\BibitemShut {NoStop}%
\bibitem [{\citenamefont {Li}\ \emph {et~al.}(2021)\citenamefont {Li},
  \citenamefont {Xu}, \citenamefont {Zhang}, \citenamefont {Shu},\ and\
  \citenamefont {Li}}]{li2021development}%
  \BibitemOpen
  \bibfield  {author} {\bibinfo {author} {\bibfnamefont {Z.}~\bibnamefont
  {Li}}, \bibinfo {author} {\bibfnamefont {H.}~\bibnamefont {Xu}}, \bibinfo
  {author} {\bibfnamefont {X.}~\bibnamefont {Zhang}}, \bibinfo {author}
  {\bibfnamefont {F.}~\bibnamefont {Shu}}, \ and\ \bibinfo {author}
  {\bibfnamefont {Z.}~\bibnamefont {Li}},\ }\bibfield  {title} {\enquote
  {\bibinfo {title} {Development of a modified {Bouc-Wen} model for butterfly
  hysteresis behaviors},}\ }in\ \href@noop {} {\emph {\bibinfo {booktitle}
  {2021 IEEE 7th International Conference on Control Science and Systems
  Engineering (ICCSSE)}}}\ (\bibinfo {organization} {IEEE},\ \bibinfo {year}
  {2021})\ pp.\ \bibinfo {pages} {159--163}\BibitemShut {NoStop}%
\end{thebibliography}%

\end{document}